\documentclass[journal]{IEEEtran}
\IEEEoverridecommandlockouts
\usepackage{lipsum}
\usepackage{hyperref}
\usepackage{placeins}
\usepackage{mathtools}
\usepackage{graphicx}
\usepackage{float}
\usepackage{setspace}
\usepackage{color}
\usepackage[dvipsnames]{xcolor}
\usepackage{cite}
\usepackage{indentfirst}
\usepackage{amssymb}
\usepackage{amsmath}
\usepackage{amsbsy}
\usepackage{subfigure}
\usepackage{graphicx}
\usepackage{setspace}
\usepackage{soul}

\begin{document}

\title{Mapping Human Muscle Force to Supernumerary Robotics Device for Overhead Task Assistance}

\author{Jianwen Luo, Sicong Liu, Chengyu Lin, Yong Zhou, Zixuan Fan, Zheng Wang, Chaoyang Song, H. Harry Asada, Chenglong Fu
\thanks{This work was supported in part by National Natural Science Foundation of China under Grant 51905251, and Centers for Mechanical Engineering Research and Education at MIT and SUSTech. \textit{(Corresponding author: C. Fu. Email: fucl@sustech.edu.cn, J. Luo. Email: jamesluo@cuhk.edu.cn)}}
\thanks{J. Luo is with Centers for Mechanical Engineering Research and Education at MIT and SUSTech; also with Shenzhen Institute of Artificial Intelligence and Robotics for Society (AIRS), Shenzhen 518172, China; also with Institute of Robotics and Intelligent Manufacturing (IRIM), The Chinese University of Hong Kong (CUHK), Shenzhen 518172, China.}
\thanks{J. Luo, S. Liu, C. Lin, Y. Zhou, Z. Fan, Z. Wang, C. Song and C. Fu are with the Department
of Mechanical Engineering, Southern University of Science and Technology, China. They are also with Centers for Mechanical Engineering Research and Education at MIT and SUSTech.}
\thanks{H. Harry Asada is with the Department of Mechanical Engineering, Massachusetts Institute of Technology, USA.}
}
\markboth{IEEE/ASME AIM 2020 Workshop on Supernumerary Robotic Devices}%
{Shell \MakeLowercase{\textit{et al.}}: Human-SuperLimb communication for Human-Centered Overhead Task Assistance}

\maketitle

\begin{abstract}
Supernumerary Robotics Device (SRD) or Supernumerary Robotic Limbs (SRL) is an ideal solution to provide robotic assistance in overhead manual manipulation, especially the tasks including supporting a panel and fitting it in the ceiling in limited space such as compartment. Since two arms are occupied for the overhead task, it is desired to have additional arms to assist us in achieving other subtasks such as supporting the far end of a long panel and actively pushing it upward to fit in the ceiling. 
In this study, a method that maps human muscle force to SRD for overhead task assistance is explored. Our idea is to utilize redundant DoFs such as the idle muscles in the leg to control the supporting force of the SRD. A sEMG device is worn on the operator’s shank where muscle signals are measured, parsed, and transmitted to SRD for control. In the control aspect, we adopted stiffness control in the task space of the SRD. The sEMG signals detected from human muscles are extracted, filtered, rectified, and parsed to estimate the muscle force. The muscle force estimated by sEMG is mapped to the desired force in the task space of the SRD. This force information is taken as the intent of the operator for proper overhead supporting force. Through tuning the stiffness and equilibrium point, the supporting force of SRD in task space can be easily controlled. The desired force is transferred into stiffness or equilibrium point to output the corresponding supporting force. 
In the preliminary test, we demonstrated the muscle force estimation using sEMG signals from shank, active stiffness control in SRD task space and supporting force control through mapping sEMG to equilibrium point of stiffness in SRD task space. A SRD prototype is integrated with a sEMG device, a 6-axis force sensor, and a visual odometry camera. Experiment results in the preliminary tests are presented to prove the feasibility of the proposed method.
\end{abstract}
\begin{IEEEkeywords}
Supernumerary robotics device, wearable robotics, overhead task, human muscle
\end{IEEEkeywords}

\section{Related work}
\IEEEPARstart{S}{upernumerary} robotic device (SRD) or Supernumerary Robotic Limbs (SRL) is an promising solution for the overhead tasks assistance in daily life or industrial field, e.g. in an aircraft's compartment \cite{parietti2014supernumerary}, due to its ability to augment the operator's manipulation 
\begin{figure}[t]
\centering
\setlength{\abovecaptionskip}{-0.3cm}
\includegraphics[width=3.5in]{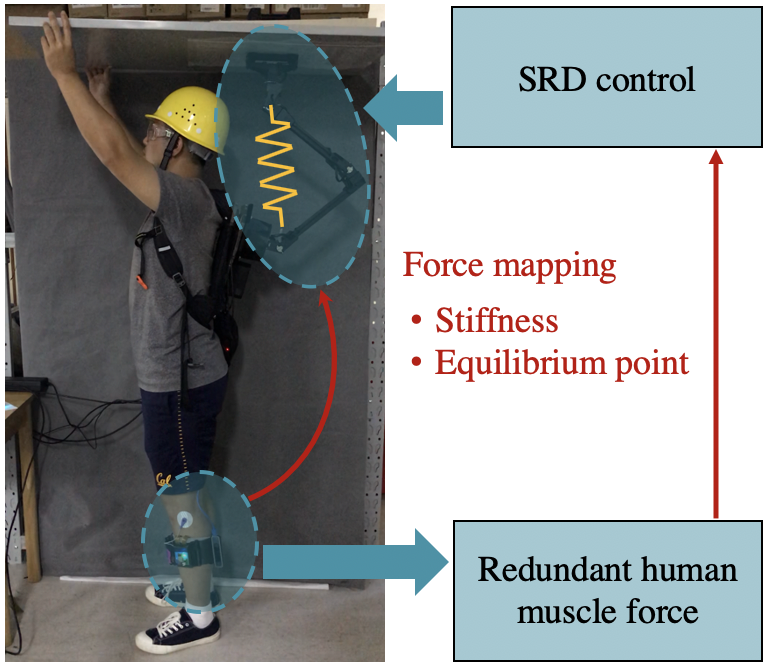}
\caption{A single operator is holding a long board with a SRD assisting in supporting it. The operator wears a sEMG device on the left shank. There is a motion sensor mounted on the sEMG device to detect the rotation angle. A F/T sensor is mounted on SRD for measuring the interacting force between operator and SRD.}
\label{fig:SRD_method}
\end{figure}
capability and extend the work space that is not constrained by the human limb movement. Some researchers pioneered in overhead-task-oriented SRD and achieved fruitful results. B. L. Bonilla and H. H. Asada designed a shoulder SRD for a single operator, which otherwise needs the collaboration between two operators \cite{bonilla2014robot}. They proposed a coloured petri nets hybrid controller to address the coordination and communication between SRL and the operator in a ceiling panel installation task. Z. Bright and H. H. Asada adopted admittance control to enable the SRL to be capable of securely holding an object or supporting the ceiling while the operator's body can move freely \cite{bright2017supernumerary}. In their study, velocity-based control in the joint level was adopted. J. Whitman and H. Choset proposed an optimal design method for SRL in the overhead tasks \cite{8594657}. However, actively controlling the supporting forces of SRD in overhead tasks, such as lifting a panel higher or fitting it in the ceiling, has not been explored yet. In contrast with velocity control in \cite{bright2017supernumerary}, utilizing the torque interface allows compliant interaction with the environment and humans residing in the work space of the robot \cite{dietrich2012reactive}. 
In this study, active stiffness control using muscle force mapping for SRL in overhead lifting and pushing is adopted.

\begin{figure}[h]
\centering
\setlength{\abovecaptionskip}{-0.1cm}
\includegraphics[width=3.0in]{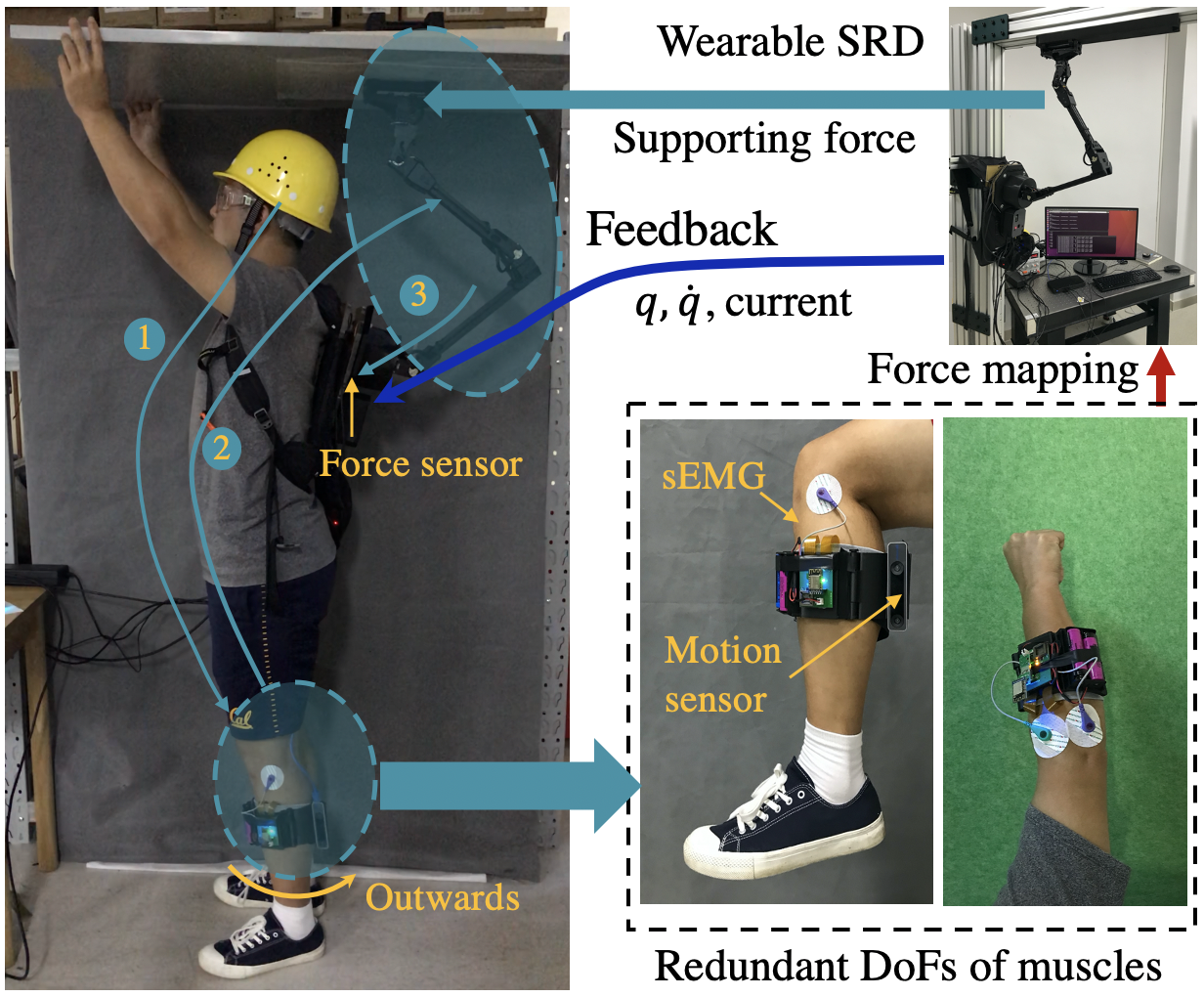}
\caption{The framework for communication between the operator and SRD. The sEMG device is integrated with visual odometry camera as the motion sensor.}
\label{fig:framework}
\end{figure}
\vspace{-5mm}
\section{Method and preliminary results}
The overall framework is shown in Fig. \ref{fig:framework}. The force of gastrocnemius muscle in the operator's leg or arm is estimated. The SRD is controlled to imitate the behaviors of muscle and output supporting forces to accomplish the overhead lifting and pushing task. We are motivated by the fact that humans can achieve daily manipulation merely through simple inherent compliance property in joint driven by muscles. As one of the well-known compliance control methods, stiffness control is easy to achieve using a few of straightforward parameters such as stiffness and equilibrium point. Therefore, stiffness and equilibrium point are used as two control parameters. Active supporting force in SDR task space is controlled through varying the stiffness or equilibrium point of the virtual spring in SRD task space. Hill model is used to parse the sEMG to identify the muscle force. There are extensive studies on muscle stiffness or force using sEMG such as in \cite{5975474, 8320381}. We made exploration using sEMG in a similar method to map muscle stiffness and force to the SRD. The research is still undergoing and some preliminary results are presented below.

\begin{figure}[h]
\centering
\setlength{\abovecaptionskip}{-0.1cm}
\includegraphics[width=2.6in]{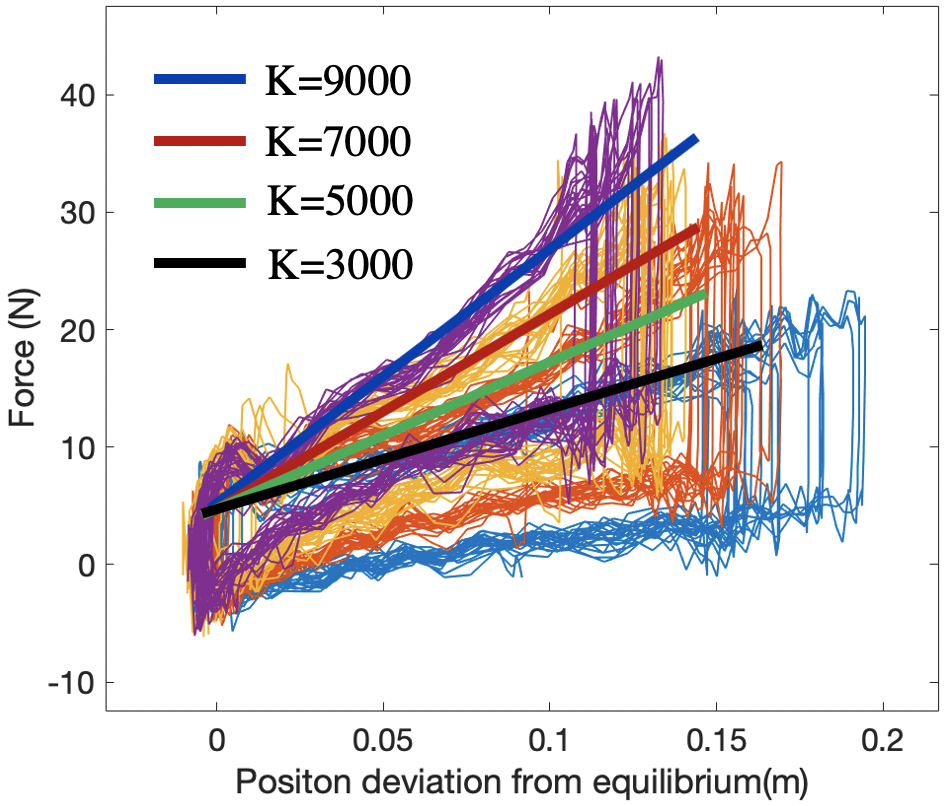}
\caption{Active stiffness control in SRD task space. Four levels of stiffness are set and presented using four colors. The straight lines are the desired stiffness.}
\label{fig:stiffness}
\end{figure}

Stiffness control is tested in SRD's task space. Fig. \ref{fig:stiffness} demonstrates the preliminary results for various stiffness in SRD task space. On one hand, setting stiffness makes SRD complaint, which is safe for the operator and stable for control; on the other hand, when SRD is detached from the panel or the weight of panel varies due to unexpected disturbance, stiffness property is able to adapt to the disturbances. Four levels of stiffness are set as desired stiffness in the tests. There exists obvious hysteresis due to the frictions in joint's actuators as can be seen in Fig. \ref{fig:stiffness}.

\begin{figure}[h]
\centering
\setlength{\abovecaptionskip}{-0.1cm}
\includegraphics[width=3.3in]{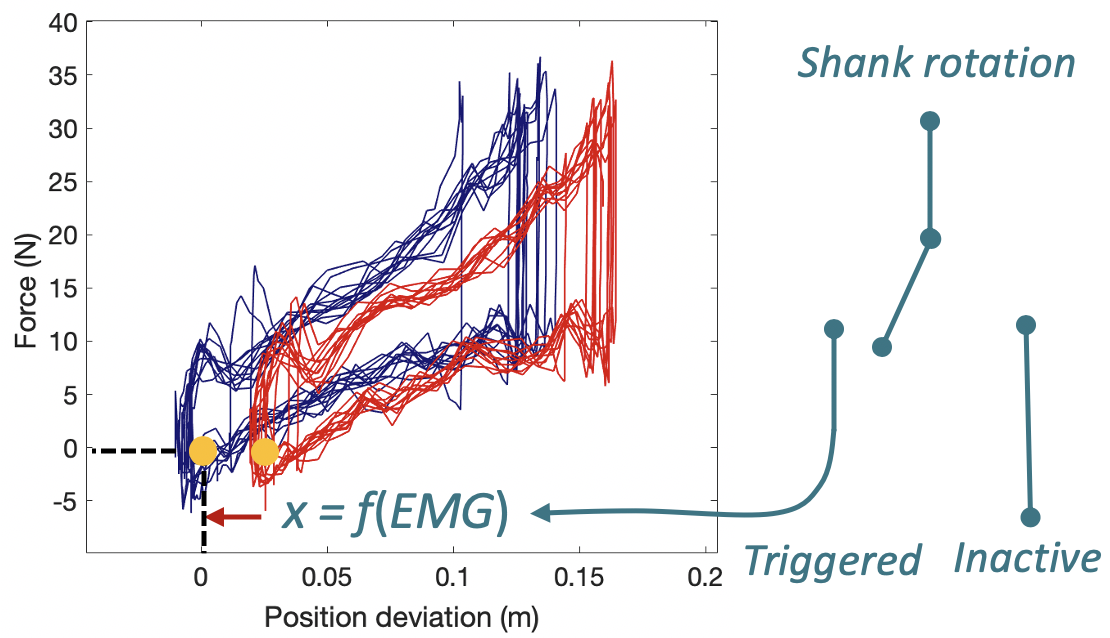}
\caption{Control the supporting force through change the equilibrium point of the task space stiffness. Motion sensor is utilized to trigger the state of sEMG mapping. When the operator's leg equipped with the sEMG device turns outward, motion sensor can measure the rotation around vertical axis and then sEMG signals are taken as active for mapping.}
\label{fig:equ_pt}
\end{figure}

Fig. \ref{fig:equ_pt} demonstrates the preliminary results for supporting force control through changing the equilibrium point. Due to the noises in sEMG signals, a visual odometry camera is integrated with the sEMG device and fused with EMG information as shown in Fig. \ref{fig:framework}. Visual odometry camera outputs stable spatial information. We utilized the rotation movement of shank wearing the integrated sensors to set a threshold for sEMG. When left foot turns outward, sEMG mapping is enabled to vary the equilibrium point towards upward. Then SRD will output larger supporting force proportional to the position deviation between new and original equilibrium point. This force pushes the panel upward. 

\vspace{-5mm}

\vspace{-5mm}
\bibliographystyle{IEEEtran}
\bibliography{mybib}
\end{document}